\documentclass[runningheads]{llncs}
\usepackage[T1]{fontenc}
\usepackage{graphicx}
\usepackage[table]{xcolor}
\usepackage{colortbl}
\usepackage{multirow}
\usepackage{collcell}

\usepackage{latexsym}
\usepackage{colortbl}

\usepackage[T1]{fontenc}

\usepackage[utf8]{inputenc}

\usepackage{microtype}

\usepackage{hhline}
\usepackage{booktabs}
\usepackage{xspace}
\usepackage{makecell}
\usepackage{vcell}
\usepackage{xcolor}
\usepackage{float}
\usepackage[nocompress]{cite}

\usepackage[normalem]{ulem}
\useunder{\uline}{\ul}{}

\usepackage[colorlinks=true]{hyperref}
\usepackage{color}
\hypersetup{
    linkcolor=blue,  
    citecolor=blue,  
    urlcolor=blue    
}

\begin{document}

\newcommand{\ha}[1]{%
    \cellcolor[rgb]{%
        \fpeval{(#1)/100 > 0.5 ? 0.95 - 0.55*((#1)/100-0.5) : 0.95},
        \fpeval{(#1)/100 > 0.5 ? 0.90 : 0.60 + 0.60*(#1)/100},
        \fpeval{0.55}}#1%
}

\newcommand{\modelname}{AMALIA-VL}

\newcommand*{\diogo}[1]{{\textcolor{red}{[DS: #1]}}}
\title{\modelname: A Native European Portuguese \\ Open-Source Vision and Language Model}

\titlerunning{\modelname}

\authorrunning{D. Glória-Silva et al.}

\author{
 Diogo Glória-Silva\textsuperscript{1,2},
 João Cardeira\textsuperscript{1,2},
 Manuel Letras da Luz\textsuperscript{1},
\\
 Afonso Simplício\textsuperscript{1,2},
 Gonçalo Vinagre\textsuperscript{1,2},
 Diogo Tavares\textsuperscript{1,2},
 Rafael Ferreira\textsuperscript{1,2},
\\
 Inês Calvo\textsuperscript{1},
 Inês Vieira\textsuperscript{1},
 David Semedo\textsuperscript{1,2},
 João Magalhães\textsuperscript{1,2}
\vspace{6pt}
}

\institute{
\textsuperscript{1}NOVA School of Science and Technology,
 \textsuperscript{2}NOVA LINCS
 \\
 \textbf{Correspondence:} \href{mailto:dmgc.silva@fct.unl.pt}{dmgc.silva@fct.unl.pt}}
 
\maketitle              
\begin{abstract}

Large Vision and Language Models (LVLMs) have advanced rapidly, yet European Portuguese (pt-PT) remains systematically underserved by existing open-source multimodal models, which either conflate it with Brazilian Portuguese or severely under-represent it in their training data mixes. We introduce \modelname{}, the first open-source instruction-tuned LVLM built natively for pt-PT, pairing a high-resolution vision encoder with dynamic image tiling and a fully open pt-PT-optimized language model via a learned connector. We contribute with a purposefully designed three-stage training process — vision-language alignment, general visual instruction tuning, and preference optimization — together with a pt-PT-centric multimodal data mix combining curated and translated public datasets with novel datasets that address the near-total absence of European Portuguese multimodal resources.
Our evaluation shows that \modelname{} establishes a strong baseline for open-source pt-PT LVLMs.
We will release model weights, training data, and construction pipelines along with machine-translated pt-PT evaluation benchmarks to help democratize pt-PT LVLM development.

\end{abstract}

\section{Introduction}

\begin{figure*}[t]
  \centering
  \includegraphics[trim={10 15 10 0},clip,width=\textwidth]{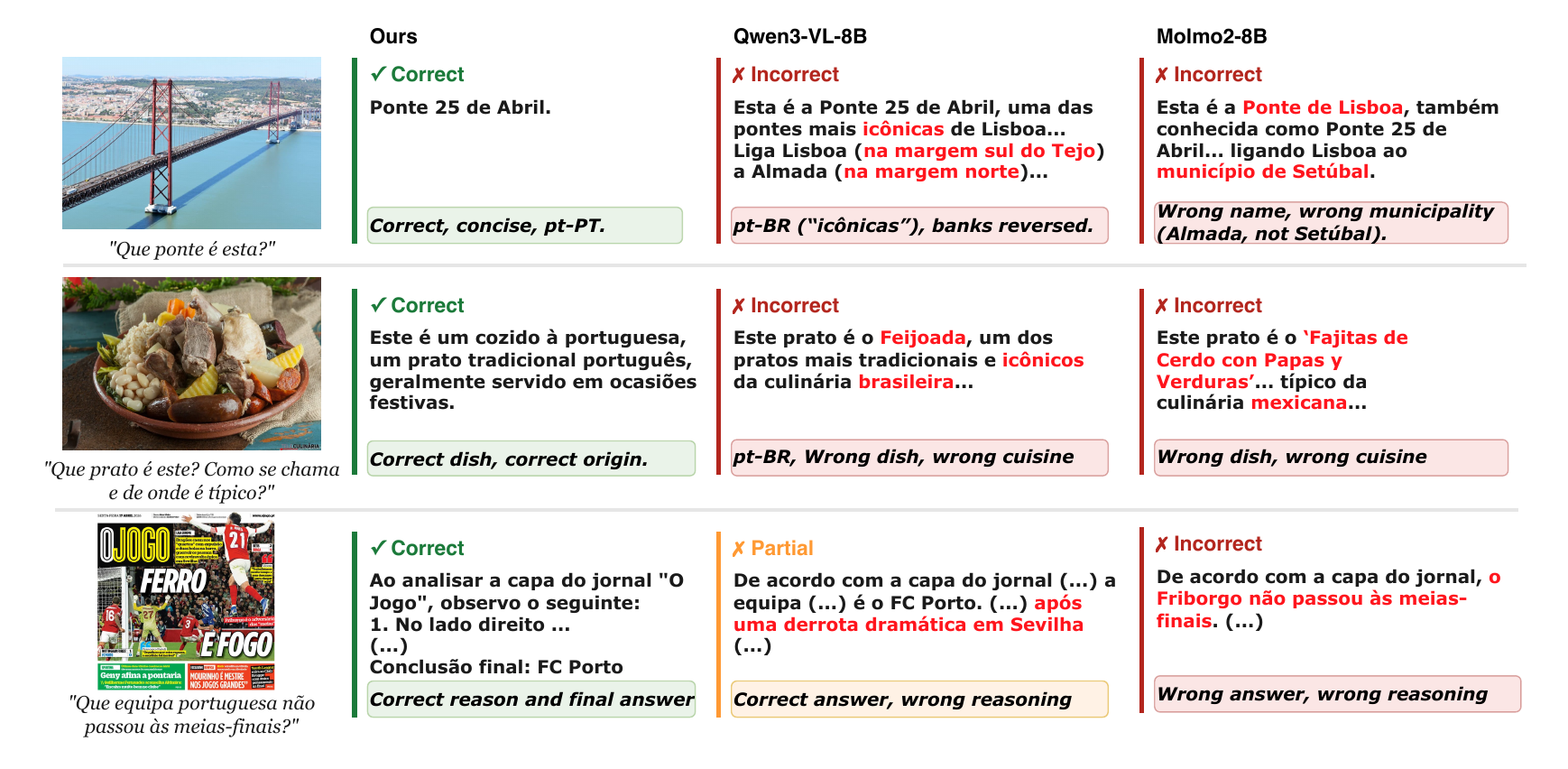}
  \vspace{-20pt}
    \caption{\modelname{} is natively European Portuguese grounding its answers in Portuguese visual culture, whereas general LVLMs hallucinate or fall back to Brazilian Portuguese.}
  \label{fig:teaser}
\end{figure*}

Large Vision and Language Models (LVLMs) have made remarkable strides in multimodal reasoning~\cite{qwen3_5, gemma3, internvl_3_5}, but their linguistic coverage remains skewed toward high-resource languages~\cite{alba}. European Portuguese (pt-PT) is a compelling case, as, despite having more than 10 million native speakers, it is consistently overshadowed by Brazilian Portuguese (pt-BR) in web-scale training corpora. Consequently, multilingual and multimodal models systematically underperform in pt-PT tasks~\cite{amalia}, exhibiting a strong bias towards the pt-BR variant and a suboptimal representation of the lexical, grammatical, and cultural conventions of pt-PT (see Figure~\ref{fig:teaser}). The closest prior work — V-GlórIA~\cite{vgloria} and TowerVision~\cite{towervision} — address Portuguese and broad European multilingualism respectively, yet neither is designed natively for pt-PT.

This creates a two pronged challenge: models lack the multimodal capabilities to process pt-PT accurately, and the community lacks the benchmarks to measure pt-PT multimodal capabilities, as, to the best of our knowledge, no multimodal evaluation resources exist for European Portuguese. 

To address both gaps, we introduce \modelname{}, the first open-source LVLM built natively for pt-PT alongside a suite of translated multimodal benchmarks. The architecture of \modelname{} follows LLaVA-NeXT~\cite{llavanext} supporting dynamic image tiling for high-resolution input and consists of a vision encoder, a modality connector, and a fully open language model targeting the European Portuguese language variant. 

Our main contributions are: \textbf{(i)} \modelname{}, the first fully open native pt-PT LVLM, competitive with leading open-source LVLMs on pt-PT evaluations; \textbf{(ii)} a three-stage multimodal training process designed to progressively instil vision capabilities while aiming to preserve the base LLM's pt-PT proficiency and cultural knowledge; \textbf{(iii)} a pt-PT-centric multimodal data mix combining high-quality public datasets with several novel synthetic datasets that fill the near-total absence of European Portuguese multimodal resources; and \textbf{(iv)} machine-translated pt-PT vision benchmarks enabling rigorous vision-based language-targeted evaluation on wide array of multimodal tasks.
Model weights, all training data, training code, and benchmark translations will be publicly released. 

\section{Related Work}

LVLM progress has been marked by a tension between capability and transparency. State-of-the-art proprietary models keep weights, data, and implementation details private, prompting a growing body of open alternatives at various degrees of transparency. We analyze this spectrum to position \modelname{}.
On one end, we have open-weights models, such as Qwen~\cite{qwen3_5}, Gemma~\cite{gemma3} and GLM~\cite{glm}, which release model weights but keep both training data and key details private. More transparent models build on open-weights LLM backbones but release vision-language training data~\cite{llava_ov_1_5}, enabling reproducible multimodal training while the language foundation remains unreproducible. On the opposite end, open-source models provide a fully reproducible and transparent pipeline combining an open-source LLM, open vision-language training data, and open weights~\cite{molmo2, salamandra, perceptionlm}. 

Across this entire spectrum of openness, European Portuguese (pt-PT) remains severely underserved at all tiers of openness, with no instruction-following LVLMs directly targeting it. Given the Brazilian Portuguese bias in web-scale corpora, this imbalance~\cite{alba} propagates from language-only pre-training into multimodal models that build on top of these foundations. Even in multilingual models~\cite{eurollm, towervision}, pt-PT is not a central pillar, falling victim to the same multilingual compromise that weakens its performance in other generalist models. V-GlórIA~\cite{vgloria} specifically targets pt-PT but lacks instruction-following capabilities, hindering its impact. 
To address this critical gap, we introduce \modelname{}, the first native pt-PT open instruction-following LVLM.

\section{Model Architecture}
To design \modelname{}, we built on previous research on LVLMs~\cite{llavanext} and paired a vision encoder with language decoder via a connector module. While the language decoder serves as the core knowledge and language understanding foundation, the vision encoder extracts semantic image features, which are then projected to the language decoder input subspace through the connector. 
Specifically, we followed the architecture proposed with LLaVA-NeXT~\cite{llavanext}, and used as vision encoder SigLip2-SO400M-patch16-384~\cite{siglip2}, which provides a good balance between performance and vision token count. 
To support high-resolution inputs, adaptive tiling was adopted, in which each input image is partitioned into a grid of tiles --- selected to best match the image's aspect ratio --- alongside a downsampled thumbnail that preserves global context. Both tiles and thumbnail are encoded by the same shared vision encoder.
For the LLM, we used the DPO version of AMALIA~\cite{amalia}, as it provides a strong and open European Portuguese-centric base with instruction following capabilities. The connector module was defined as a two-layer MLP with a GELU activation.  
Different connector configurations were tested, including a linear, Q-former, among others, but the MLP yielded the best results.

\section{\modelname~Training Process}

In complex, multiple neural module networks, multi-stage training emerges as a key methodology for training convergence~\cite{llava_ov_1_5, nemotronv2}. In ~\modelname{}, we followed a multi-stage training approach that progressively instils vision capabilities while seeking to minimize text-only instruction following capability regression in pt-PT. 
In this section, we provide a detailed overview of each training stage, its motivation, and datasets used. 

\subsection{Stage 1: Vision-Language Alignment}
Our initial training stage followed~\cite{llava_ov_1_5} and focuses on vision-language alignment. Specifically, this stage served as the warmup for the connector module, initializing it to a stable foundation for vision to text alignment. To achieve this, we froze the vision encoder and the language decoder, and trained solely the connector module on image captioning data using 500k samples from the PD12M~\cite{pd12m} dataset, a large scale image-text open domain dataset. For this stage, we disabled tiling.

\subsection{Stage 2: General Visual Instruction Tuning}

\begin{table*}[!t]
\centering
\caption{\modelname{}'s Stage 2 Training data mixture. $\dagger$ denotes in-house synthetic datasets. }
\label{tab:sft_mix}
\resizebox{\textwidth}{!}{%
\begin{tabular}{@{} l l l l l l @{}}
\hline
\colorbox[rgb]{0.92,0.86,0.80}{\textbf{Grounding (30.4\%)}} & \textcolor[rgb]{0.92,0.86,0.80}{\rule{1.2ex}{1.2ex}} Nemotron2 OI BBox1(500K) & \textcolor[rgb]{0.92,0.86,0.80}{\rule{1.2ex}{1.2ex}} Nemotron2 OI BBox2(500K) & \\
& \textcolor[rgb]{0.92,0.86,0.80}{\rule{1.2ex}{1.2ex}} Nemotron2 OI BBox3(500k) & \textcolor[rgb]{0.92,0.86,0.80}{\rule{1.2ex}{1.2ex}} TallyQA(98.7K) & \\
\hline
\colorbox[rgb]{0.98,0.85,0.87}{\textbf{General VQA (19.7\%)}} & \textcolor[rgb]{0.98,0.85,0.87}{\rule{1.2ex}{1.2ex}} PT-VQA-Gen$^\dagger$(543K) & \textcolor[rgb]{0.98,0.85,0.87}{\rule{1.2ex}{1.2ex}} MMEvol(157K) & \textcolor[rgb]{0.98,0.85,0.87}{\rule{1.2ex}{1.2ex}} VisDial(123K) &  & \\ & \textcolor[rgb]{0.98,0.85,0.87}{\rule{1.2ex}{1.2ex}} VQAv2(82.8K) &
\textcolor[rgb]{0.98,0.85,0.87}{\rule{1.2ex}{1.2ex}} LLaVA-150K(81.5K) & \textcolor[rgb]{0.98,0.85,0.87}{\rule{1.2ex}{1.2ex}} Nemotron VQA9(46.7K) \\
\hline
\colorbox[rgb]{0.84,0.93,0.97}{\textbf{Naive OCR (11.8\%)}} & \textcolor[rgb]{0.84,0.93,0.97}{\rule{1.2ex}{1.2ex}} Nemotron OCR4-5(382K) & \textcolor[rgb]{0.84,0.93,0.97}{\rule{1.2ex}{1.2ex}} SimpleCodeOCR$^\dagger$(175K) & \textcolor[rgb]{0.84,0.93,0.97}{\rule{1.2ex}{1.2ex}} Nemotron OCR2(29.1K) & \\ & \textcolor[rgb]{0.84,0.93,0.97}{\rule{1.2ex}{1.2ex}} Nemotron OCR1(14.5K) &
\textcolor[rgb]{0.84,0.93,0.97}{\rule{1.2ex}{1.2ex}} Nemotron OCR3(14.5K) & \textcolor[rgb]{0.84,0.93,0.97}{\rule{1.2ex}{1.2ex}} IIIT5K(2.0K) \\
\hline
\colorbox[rgb]{0.90,0.88,0.96}{\textbf{Text IF (9.6\%)}} & \textcolor[rgb]{0.90,0.88,0.96}{\rule{1.2ex}{1.2ex}}AMALIA-LLM SFT Data (500k) & \textcolor[rgb]{0.90,0.88,0.96}{\rule{1.2ex}{1.2ex}} Persona Nemotron(4.5K) & \textcolor[rgb]{0.90,0.88,0.96}{\rule{1.2ex}{1.2ex}} Self Identification (156) &  \\
\hline
\colorbox[rgb]{0.82,0.94,0.90}{\textbf{Chart \& Table (7.1\%)}} & \textcolor[rgb]{0.82,0.94,0.90}{\rule{1.2ex}{1.2ex}} Nemotron OCR9(224K) & \textcolor[rgb]{0.82,0.94,0.90}{\rule{1.2ex}{1.2ex}} InfographicSynth$^\dagger$(96K) & \textcolor[rgb]{0.82,0.94,0.90}{\rule{1.2ex}{1.2ex}} Nemotron VQA4-7-8(53.7K) &  \\
\hline
\colorbox[rgb]{0.90,0.90,0.90}{\textbf{Captioning (6.7\%)}} & \textcolor[rgb]{0.90,0.90,0.90}{\rule{1.2ex}{1.2ex}} PT-Caps$^\dagger$(250K) & \textcolor[rgb]{0.90,0.90,0.90}{\rule{1.2ex}{1.2ex}} PT-Caps-Fusion$^\dagger$(100K) &  &  \\
\hline
\colorbox[rgb]{0.99,0.89,0.81}{\textbf{OCR QA (5.9\%)}} & \textcolor[rgb]{0.99,0.89,0.81}{\rule{1.2ex}{1.2ex}} OCR-VQA(166K) & \textcolor[rgb]{0.99,0.89,0.81}{\rule{1.2ex}{1.2ex}} PT-OCR$^\dagger$(71.3K) & & \\
 & \textcolor[rgb]{0.99,0.89,0.81}{\rule{1.2ex}{1.2ex}} PT-Render-Text$^\dagger$(50K) & \textcolor[rgb]{0.99,0.89,0.81}{\rule{1.2ex}{1.2ex}} TextVQA(21.8K) & \\
\hline
\colorbox[rgb]{0.85,0.85,0.95}{\textbf{Code Reasoning (3.7\%)}} & \textcolor[rgb]{0.85,0.85,0.95}{\rule{1.2ex}{1.2ex}} PTSimpleCodeOutputs$^\dagger$(117K) & \textcolor[rgb]{0.85,0.85,0.95}{\rule{1.2ex}{1.2ex}} SePIC$^\dagger$(44.3K) & \textcolor[rgb]{0.85,0.85,0.95}{\rule{1.2ex}{1.2ex}} PTOutputsToCode$^\dagger$(34.6K) &  \\
\hline
\colorbox[rgb]{0.93,0.84,0.94}{\textbf{Mathematics (2.3\%)}} & \textcolor[rgb]{0.93,0.84,0.94}{\rule{1.2ex}{1.2ex}} CLEVR-Math(70K) & \textcolor[rgb]{0.93,0.84,0.94}{\rule{1.2ex}{1.2ex}} CoSyn-400K Math(40.7K) & \textcolor[rgb]{0.93,0.84,0.94}{\rule{1.2ex}{1.2ex}} Geomverse(8.6K) &  \\
\hline
\colorbox[rgb]{0.82,0.94,0.95}{\textbf{Culture (1.8\%)}} & \textcolor[rgb]{0.82,0.94,0.95}{\rule{1.2ex}{1.2ex}} Caravela SFT$^\dagger$ (96.8K) &  &  &  \\
\hline
\colorbox[rgb]{0.92,0.96,0.82}{\textbf{Doc. Understanding (0.9\%)}} & \textcolor[rgb]{0.92,0.96,0.82}{\rule{1.2ex}{1.2ex}} Nemotron2-DocVQA-CoT(36.3K) & \textcolor[rgb]{0.92,0.96,0.82}{\rule{1.2ex}{1.2ex}} InvoiceQA$^\dagger$(8.6K) &  &  \\
\hline
\colorbox[rgb]{0.99,0.96,0.82}{\textbf{Science (0.1\%)}} & \textcolor[rgb]{0.99,0.96,0.82}{\rule{1.2ex}{1.2ex}} AI2D(4.9K) & & &  \\
\hline
\end{tabular}
}
\end{table*}

The second training stage was the largest and targeted full model training on a highly diverse set of visual instruction-following tasks, fine-tuning \modelname{} for fine-grained image comprehension. 
In this stage, the model learned to selectively attend to visual inputs depending on the provided instruction. 

To gather comprehensive visual instruction-tuning data, we used open-license datasets and complemented them with targeted synthetic datasets (see \S\ref{sec:synth_data}) that were designed to elicit specific behaviours absent from the collected data and to address the scarcity of pt-PT resources. As illustrated in Table~\ref{tab:sft_mix}, our data mix of 4.7M samples, totalling $\approx$2B text tokens, covered the following multimodal tasks:
\textbf{Grounding}~\cite{nemotronv2, tallyqa},
\textbf{General VQA}~\cite{mmevol, visdial, vqav2, llavanext, eagle2}, 
\textbf{Naive OCR}~\cite{eagle2, iiit5k},
\textbf{Chart \& Table}~\cite{eagle2},
\textbf{Captioning},
\textbf{OCR QA}~\cite{ocr_vqa, textvqa},
\textbf{Code Reasoning},
\textbf{Mathematics}~\cite{clevr_math, cosyn, geomverse},
\textbf{Document Understanding}~\cite{nemotronv2, fatura},
\textbf{Text Instruction Following}~\cite{amalia},
and \textbf{Science}~\cite{ai2d, scienceqa}.

To extend pt-PT coverage, we aimed to eliminate what we call "monolingual islands", where a task's monolingual presence in training hinders multilingual performance transfer. We tackled this by translating several datasets using a combination of TranslateGemma~\cite{translategemma} and Gemma3~\cite{gemma3}, which were selected due to their strong pt-PT proficiency in EuroEval~\cite{smart2024encoder}. 

\subsection{Stage 3: Preference Optimization}
This stage used Direct Preference Optimization (DPO)~\cite{dpo} and sought to increase the model's likelihood of generating preferred responses while minimizing undesirable patterns.
Due to the lack of publicly available multimodal preference optimization datasets, we relied on automated synthetic preference annotations derived from the Stage 2 data mix, coupled with answer rewriting, as detailed in \S\ref{sec:synth_data_dpo}. Sampling from this mix ensures not only that we stay within the model distribution, but also that all task types are covered during preference optimization. Additionally, we incorporated 100k samples from InternVL3.5's~\cite{internvl_3_5} MPO dataset, excluding any samples derived from sources with restrictive licenses. And finalize the mix with safety alignment data.

\section{Synthetic Dataset Creation}
\label{sec:synth_data}

Developing an open-source, transparent, and viable pt-PT LVLM has three primary data requirements: \textbf{(1)} open licenses, and not just research-only, vision-language datasets, \textbf{(2)} publicly available pt-PT VQA, instruction tuning, or OCR resources, and \textbf{(3)} VQA annotations of raw large scale public-domain image collections.
To tackle these requirements, we generated targeted synthetic datasets using open-source models and public-domain image collections. This allowed for native and task-specific pt-PT support. Next, we detail the synthetic pipeline for creating both visual instruction tuning (Stage 2) and preference optimization data (Stage 3).

\subsection{General Visual Instruction Tuning Data Mixture}

\begin{figure*}[t]
  \centering
  \includegraphics[trim={20 16 20 12},clip,width=\textwidth]{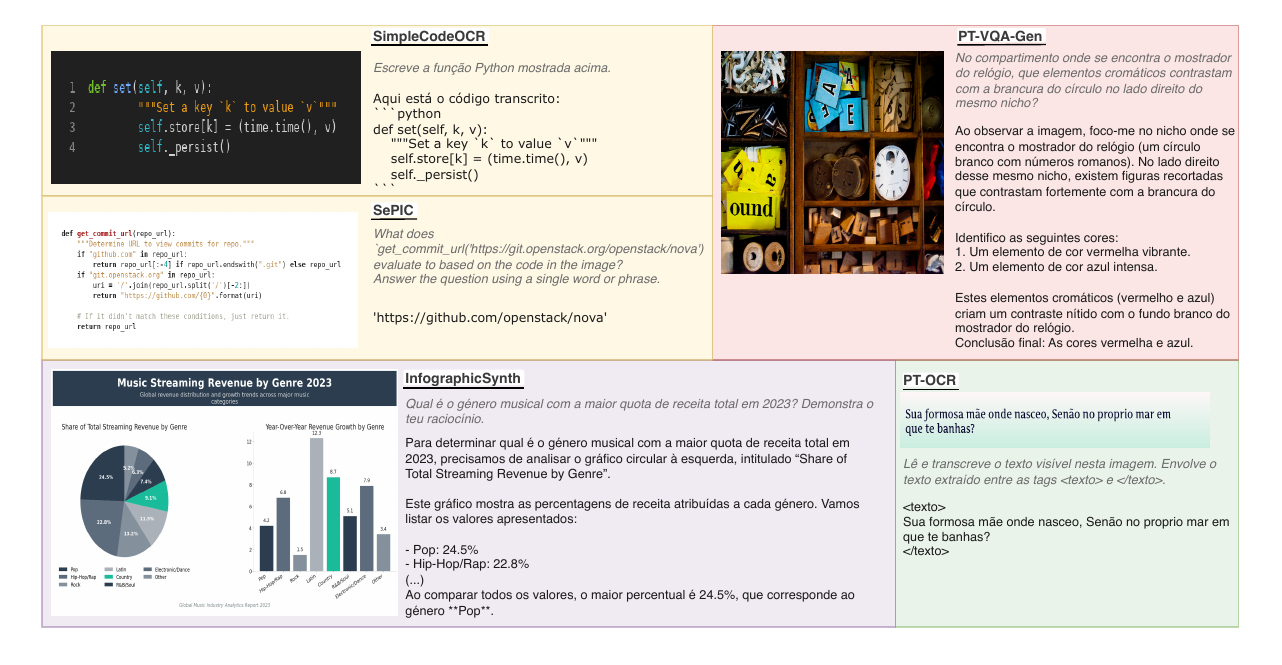}
  \vspace{-20pt}
    \caption{Samples from several of our pt-PT focused synthetic datasets.}
  \label{fig:infovqa_ptocr_examples}
  \label{fig:code_vqa_examples}
  ~\vspace{-20pt}
\end{figure*}

\paragraph{\textbf{PT-OCR.}} This dataset targets native European Portuguese OCR training data with a strong emphasis on adherence to diverse output formatting conventions and formats, see Figure~\ref{fig:infovqa_ptocr_examples}.
PT-OCR combines an annotated pt-PT OCR dataset~\cite{mazaafard} as seed with a template-based approach to create OCR-centric dialogues, targeting 3 tasks: naive OCR (verbatim transcription of the full text), sentence-level extraction (first, last, $n$-th sentence, and sentence count), and hallucination detection (accept or reject a candidate transcription, providing the correct text when refuting). 
For diversity, several output formats, image perturbations, and concatenated samples were generated to produce more complex visual inputs.

\paragraph{\textbf{InvoiceQA.}}
This is an invoice-style document processing task that leverages FATURA~\cite{fatura}, a public corpus of synthetic invoices for field extraction (e.g. date, buyer name, seller name, invoice number) and rejection of incorrect field/region associations. 
Each invoice mixes two task formats: field extraction and bounding box prediction. In the former, the model is asked in natural language for a field's value, with Qwen3~\cite{qwen3} paraphrased questions, and it also rewrites the raw OCR span into a fluent answer. In the latter, given a form field, it needs to produce the correct bounding box.
We also added negative samples ($p=0.3$) pairing $A$'s field name with $B$'s bounding box. The reference answer rejects the association and identifies $B$.

\paragraph{\textbf{PT-Caps \& PT-Caps-Fusion.}}

These two datasets target bilingual image captioning at varying detail levels to improve image understanding and steerability. PT-Caps consists of 250k PD12M~\cite{pd12m} images captioned by Gemma3-27b~\cite{gemma3} and each sample has 3 levels of verbosity (small, medium and detailed) in both English and pt-PT. PT-Caps-Fusion focuses solely on descriptive captions generated by multiple LVLMs to ensure diversity. For each image, we computed the similarity of each caption against all others, excluded outliers, and fed the 3 most dissimilar captions to Qwen3VL-235B~\cite{qwen3} to generate a merged caption. The final captions were translated to pt-PT with TranslateGemma~\cite{translategemma}. 

\paragraph{\textbf{PT-VQA-Gen.}} 
This dataset introduces general-purpose natively pt-PT VQA data created via a 5-stage pipeline. First, we used MetaCLIP~\cite{xu2024demystifying} to map 5M open-domain images~\cite{pd12m, openimages, yfcc100m} to 60 curated visual mega-concepts~\cite{llava_ov_1_5}, uniformly sampling 750k. Second, to ensure diversity, an LVLM generated up to 5 vision-centric questions per image using two (out of 36 curated) visual attention personas, yielding 4.8M VQA pairs. Third, lexical and classifier-based~\cite{ptbr_classifier} filters removed pt-BR samples, leaving 3.6M. Fourth, an LVLM answered each question with a Chain-of-Thought (CoT) trace and a Gemma4-E4B verified equivalence with the original answer, discarding mismatches and hallucinations to retain 1.9M valid samples. Finally, Gemma4-E4B attempted all questions, and correctly answered (easy) samples were dropped. Surviving pairs were grouped by image into dialogues, with 15\% formatted as Multiple Choice Questions (MCQ) and 15\% as short answers without CoT. We used Gemma4-31B for generation and Gemma4-E4B~\cite{gemma4} for entailment. A sample is shown in Figure~\ref{fig:code_vqa_examples}.

\paragraph{\textbf{Code Datasets.}}

To evaluate visual code recognition and parsing, we introduce four complementary bilingual datasets (pt-PT and English).
First, to isolate pure extraction capabilities, \textbf{SimpleCodeOCR} rendered over 300k permissible CodeSearchNet~\cite{husain2019codesearchnet} functions for a naive code OCR task.
Then, \textbf{SePIC} (Semantic Parsing of Image-based Code) used $\approx$17k syntactically correct, 50-line max Python functions filtered from CodeSearchNet. These code snippets were rendered as images and paired with Gemma4-E4B generated invocations (executed in a sandbox), tasking the model to predict the execution output alongside Gemma4-31B CoT traces. Reversing this formulation, \textbf{PTOutputsToCode} rendered the execution outputs as images, formulating an MCQ task to identify the correct function-invocation pair.
Finally, \textbf{PTSimpleCodeOutputs} used Gemma4-31B to generate 100k short, executable Python snippets (max 10 lines) utilizing the \texttt{print} function. These were rendered as images for code parsing in yes/no, direct, and MCQ formats. Some samples can be seen in Figure~\ref{fig:code_vqa_examples}.

\paragraph{\textbf{InfographicSynth.}}
To target rich chart comprehension, in InfographicSynth, we assigned an LVLM, Qwen3.5-122B~\cite{qwen3_5}, to generate content for 1 of 14 curated chart templates along with candidate questions. For each sample, we generated an answer and CoT trace. Similarly to PT-VQA-Gen, we considered several output formats. Additionally, 30\% of samples are composed of 2-3 concatenated panels to increase sample complexity. To improve language transfer, we considered pt-PT and En for both the infographics and QA pairs. A sample can be seen in Figure~\ref{fig:infovqa_ptocr_examples}.

\paragraph{\textbf{CaravelaSFT.}}
This dataset provides culturally grounded SFT data for Portuguese heritage, sourced primarily from Wikipedia and the national DGADR gastronomy registry. First, we used a Qwen3.5-27B judge to filter entities for cultural relevance, retaining only the most relevant ones to construct the training corpus. Second, Gemma4-31B and Qwen3.5-27B generated diverse training samples for these entities, covering captioning, narration under three stylistic personas, multi-turn dialogues, cross-entity QA, and negative questions pairing unrelated entities. Third, deterministic regex rules filtered non-European Portuguese text, and a deduplication pass discarded near-identical entries. Finally, questions answered correctly by a blind text-only evaluator were recycled as additional training signal, and direct MCQ responses were replaced with Chain-of-Thought traces generated by Gemma4-31B to articulate step-by-step visual reasoning, favouring reasoning over memorisation. The final corpus contains 95k samples.

\subsection{Preference data}
\label{sec:synth_data_dpo}

For DPO data, we followed~\cite{amalia} and source prompts from Stage 2's data mix and generated candidates with \modelname{}. To increase candidate quality while minimizing out-of-policy candidates, we used Gemma4-31B to make small edits to \modelname{} generations to improve their quality. This led to 32 candidate answers per prompt.
For preference scoring, we found open reward models to produce inconsistent rewards for pt-PT samples, so, instead, we used an LVLM to score each candidate, specifically we used Qwen3-30B~\cite{qwen3} to avoid Gemma4 biasing to its edited answers.
We sampled 100k samples, excluding bounding box data.

To tackle safety optimization we also used safety splits from the AMALIA-LLM-DPO training~\cite{amalia}. We complement these data with three new synthetic preference datasets derived from open Not Safe For Work (NSFW) datasets:
\begin{enumerate}
    \item \href{https://huggingface.co/datasets/jjmachan/NSFW-questions}{NSFW-Questions} - A dataset of sexually charged questions and answers. We do not consider depression related questions.
    \item \href{https://huggingface.co/datasets/acheong08/nsfw_reddit}{NSFW-Reddit} - A dataset of sexual posts with a title and post body. We use the title as the user request and post body as rejected response.
    \item \href{https://huggingface.co/datasets/deepghs/nsfw_detect}{NSFW-Detect} - A collection of NSFW images spanning various topics. As it has no textual annotations we use an early version of AMALIA-VL, that was not safety tuned, to generate image descriptions. 
\end{enumerate}

For all datasets we use Gemma4-31B to generate refusals.

\section{Implementation Details}

\subsubsection*{Training Details.}

\begin{table}[t]
\centering
\caption{Hyper-parameters and dataset statistics for each training stage. Components: $V_E$ - vision encoder component, $Conn$ - connector, $LLM$ - language model.}
\label{tab:hyper}

\setlength{\tabcolsep}{4pt}
\renewcommand{\arraystretch}{1.15}

\resizebox{\textwidth}{!}{
\begin{tabular}{l >{\centering\arraybackslash}p{0.9cm} >{\centering\arraybackslash}p{3cm} >{\centering\arraybackslash}p{1.8cm} >{\centering\arraybackslash}p{1.1cm} >{\centering\arraybackslash}p{1.4cm} >{\centering\arraybackslash}p{0.9cm} >{\centering\arraybackslash}p{1.4cm} >{\centering\arraybackslash}p{0.9cm}}
\toprule
\textbf{Stage} & \textbf{Batch} & \textbf{LR \quad\quad\quad\quad\quad \small{($V_E$, $Conn$, $LLM$)}} & \textbf{Scheduler} & \textbf{Items} & \textbf{Max Seq.} & \textbf{Tiles} & \textbf{Trainable} & \textbf{GPUs} \\
\midrule
1 - Conn. Warmup & 512 & $1e^{-3}$ & Constant & 0.5M & 2048 & 1 & Conn. & 8 \\
2 - Multimodal SFT & 128 & ($2e^{-5}$, $1e^{-4}$, $2e^{-5}$) & Cosine & 4.7M & 16384 & 12 & All & 128 \\
3 - Preference Opt. & 128 & $1e^{-6}$ & Cosine & 0.3M & 16384 & 12 & All & 64 \\
\bottomrule
\end{tabular}
}
\vspace{-12pt}
\end{table}

Table~\ref{tab:hyper} has the hyper parameters used for all 3 training stages. The optimizer was AdamW~\cite{loshchilov2017decoupled}
with 0.01 weight decay, beta (0.9, 0.999), and $1e^{-8}$ epsilon. Training used DeepSpeed Zero3, with NVIDIA H100 GPUs with 64GB VRAM each, with \texttt{bfloat16} mixed precision. 

\vspace{-2mm}
\subsubsection*{Evaluation Protocol.}

As model performance is sensitive to the evaluation setup, we outlined a unified evaluation protocol that ensures fair, auditable, and standardized evaluation across tasks and models.
We used the \texttt{lmms-eval}~\cite{lmms_eval} framework and evaluated a wide suite of LVLMs across the openness spectrum~\cite{qwen3_5,qwen3,minicpmv4_5,glm,ministral3,gemma3,nemotronv2,llava_ov_1_5,towervision,eurollm,molmo2, perceptionlm,salamandra}. 
We selected instruct variants and disabled thinking for unified models~\cite{qwen3_5, glm}. 
Default model configurations were kept, except for the temperature, which we set to 0, ensuring deterministic outputs. 

We evaluate on a wide set of benchmarks across several categories including General VQA~\cite{mme, mmmu, mmstar, mmpro, seedbench, realworldqa2024, pope}, OCR and Document Understanding~\cite{ocrbench, textvqa, docvqa, infovqa}; Chart and Diagram Understanding~\cite{ai2d, chartqa}; Spatial Understanding~\cite{embsp, refcoco}; Captioning~\cite{coco, refcoco} and Mathematics~\cite{mathvision}. 
All evaluations were conducted in pt-PT. Given the scale, we relied on machine translation using Gemini-3.1-Pro, selected for its state-of-the-art multilingual performance on EuroEval~\cite{smart2024encoder} and validated through manual assessment on a representative subset.
We prompted the model to not translate names and keep mathematical notation.

\section{Results and Discussion}

\definecolor{grpGeneral}{rgb}{0.93, 0.95, 1.0}
\definecolor{grpOCR}{rgb}{1.0, 0.95, 0.90}
\definecolor{grpChart}{rgb}{0.92, 1.0, 0.92}
\definecolor{grpSpatial}{rgb}{1.0, 0.93, 0.95}
\definecolor{grpCaption}{rgb}{0.95, 0.93, 1.0}
\definecolor{grpMath}{rgb}{0.95, 1.0, 0.97}
\definecolor{grpAvg}{gray}{0.92}

\definecolor{grpGeneral}{rgb}{0.93, 0.95, 1.0}
\definecolor{grpOCR}{rgb}{1.0, 0.95, 0.90}
\definecolor{grpChart}{rgb}{0.92, 1.0, 0.92}
\definecolor{grpSpatial}{rgb}{1.0, 0.93, 0.95}
\definecolor{grpCaption}{rgb}{0.95, 0.93, 1.0}
\definecolor{grpMath}{rgb}{0.95, 1.0, 0.97}
\definecolor{grpAvg}{gray}{0.92}

\begin{table*}[!t]
    \renewcommand{\arraystretch}{1.1}
    \centering
    \caption{Results on pt-PT tasks across core V\&L tasks: VQA, OCR, Diagram and Spacial understanding, Captioning and Math.
    }
    \label{tab:main_results_portuguese}
    \small
    \resizebox{\textwidth}{!}{%
    \setlength{\tabcolsep}{4pt}
    \begin{tabular}{@{} p{4.4cm} *{19}{c} @{}}
        \toprule
         & \multicolumn{7}{c}{\textbf{General VQA}} & \multicolumn{4}{c}{\textbf{\makecell{OCR \&\\Document}}} & \multicolumn{2}{c}{\textbf{\makecell{Chart \&\\Diagram}}} & \multicolumn{2}{c}{\textbf{Spatial}} & \multicolumn{2}{c}{\textbf{Caption}} & \multicolumn{1}{c}{\textbf{Math}} & \\
        \cmidrule(lr){2-8}\cmidrule(lr){9-12}\cmidrule(lr){13-14}\cmidrule(lr){15-16}\cmidrule(lr){17-18}\cmidrule(lr){19-19}
        \textbf{Model} & \cellcolor{grpGeneral}\rotatebox{90}{\textbf{MME}\cite{mme}$^\dagger$} & \cellcolor{grpGeneral}\rotatebox{90}{\textbf{MMMU}\cite{mmmu}$^\dagger$} & \cellcolor{grpGeneral}\rotatebox{90}{\textbf{MMStar}\cite{mmstar}$^\dagger$} & \cellcolor{grpGeneral}\rotatebox{90}{\textbf{MMPro}\cite{mmpro}} & \cellcolor{grpGeneral}\rotatebox{90}{\textbf{SEED}\cite{seedbench}$^\dagger$} & \cellcolor{grpGeneral}\rotatebox{90}{\textbf{POPE}\cite{pope}$^\dagger$} & \cellcolor{grpGeneral}\rotatebox{90}{\textbf{RWQA}\cite{realworldqa2024}} & \cellcolor{grpOCR}\rotatebox{90}{\textbf{OCR}\cite{ocrbench}} & \cellcolor{grpOCR}\rotatebox{90}{\textbf{TxtVQA}\cite{textvqa}$^\dagger$} & \cellcolor{grpOCR}\rotatebox{90}{\textbf{DocVQA}\cite{docvqa}$^\dagger$} & \cellcolor{grpOCR}\rotatebox{90}{\textbf{InfoVQA}\cite{infovqa}$^\dagger$} & \cellcolor{grpChart}\rotatebox{90}{\textbf{AI2D}\cite{ai2d}} & \cellcolor{grpChart}\rotatebox{90}{\textbf{ChartQA}\cite{chartqa}} & \cellcolor{grpSpatial}\rotatebox{90}{\textbf{EmbSp}\cite{embsp}} & \cellcolor{grpSpatial}\rotatebox{90}{\textbf{RefRec}\cite{refcoco}} & \cellcolor{grpCaption}\rotatebox{90}{\textbf{COCO}\cite{coco}$^\dagger$} & \cellcolor{grpCaption}\rotatebox{90}{\textbf{RefCap}\cite{refcoco}} & \cellcolor{grpMath}\rotatebox{90}{\textbf{MVision}\cite{mathvision}$^\dagger$} & \cellcolor{grpAvg}\rotatebox{90}{\textbf{Average}} \\
        \midrule
        Ministral-3-8B \hfill $\diamondsuit$            & \cellcolor{grpGeneral}1756 & \cellcolor{grpGeneral}45.4 & \cellcolor{grpGeneral}45.3 & \cellcolor{grpGeneral}26.8 & \cellcolor{grpGeneral}70.3 & \cellcolor{grpGeneral}77.1 & \cellcolor{grpGeneral}56.2 & \cellcolor{grpOCR}61.7 & \cellcolor{grpOCR}52.8 & \cellcolor{grpOCR}70.0 & \cellcolor{grpOCR}52.7 & \cellcolor{grpChart}71.9 & \cellcolor{grpChart}49.5 & \cellcolor{grpSpatial}49.2 & \cellcolor{grpSpatial}26.9 & \cellcolor{grpCaption}22.4 & \cellcolor{grpCaption}4.0 & \cellcolor{grpMath}29.5 & \cellcolor{grpAvg}48.6 \\
        Gemma-3-12B \hfill $\diamondsuit$               & \cellcolor{grpGeneral}2120 & \cellcolor{grpGeneral}47.8 & \cellcolor{grpGeneral}51.4 & \cellcolor{grpGeneral}32.1 & \cellcolor{grpGeneral}70.5 & \cellcolor{grpGeneral}85.1 & \cellcolor{grpGeneral}49.4 & \cellcolor{grpOCR}62.7 & \cellcolor{grpOCR}61.2 & \cellcolor{grpOCR}67.8 & \cellcolor{grpOCR}42.8 & \cellcolor{grpChart}75.9 & \cellcolor{grpChart}54.2 & \cellcolor{grpSpatial}55.7 & \cellcolor{grpSpatial}6.1 & \cellcolor{grpCaption}26.7 & \cellcolor{grpCaption}5.8 & \cellcolor{grpMath}32.4 & \cellcolor{grpAvg}50.2 \\
        GLM-4.6V-Flash \hfill $\diamondsuit$            & \cellcolor{grpGeneral}2294 & \cellcolor{grpGeneral}48.6 & \cellcolor{grpGeneral}60.7 & \cellcolor{grpGeneral}36.9 & \cellcolor{grpGeneral}77.1 & \cellcolor{grpGeneral}86.8 & \cellcolor{grpGeneral}66.8 & \cellcolor{grpOCR}76.2 & \cellcolor{grpOCR}69.9 & \cellcolor{grpOCR}66.0 & \cellcolor{grpOCR}5.7 & \cellcolor{grpChart}80.6 & \cellcolor{grpChart}42.8 & \cellcolor{grpSpatial}70.6 & \cellcolor{grpSpatial}\textbf{86.9} & \cellcolor{grpCaption}25.8 & \cellcolor{grpCaption}9.8 & \cellcolor{grpMath}7.8 & \cellcolor{grpAvg}55.6 \\
        MiniCPM-V4 \hfill $\diamondsuit$                & \cellcolor{grpGeneral}2274 & \cellcolor{grpGeneral}44.9 & \cellcolor{grpGeneral}60.4 & \cellcolor{grpGeneral}27.9 & \cellcolor{grpGeneral}76.8 & \cellcolor{grpGeneral}85.7 & \cellcolor{grpGeneral}66.6 & \cellcolor{grpOCR}72.4 & \cellcolor{grpOCR}65.1 & \cellcolor{grpOCR}70.0 & \cellcolor{grpOCR}59.8 & \cellcolor{grpChart}79.0 & \cellcolor{grpChart}74.5 & \cellcolor{grpSpatial}66.2 & \cellcolor{grpSpatial}45.1 & \cellcolor{grpCaption}32.0 & \cellcolor{grpCaption}6.0 & \cellcolor{grpMath}19.1 & \cellcolor{grpAvg}57.4 \\
        Qwen3-VL-8B \hfill $\diamondsuit$               & \cellcolor{grpGeneral}\textbf{2355} & \cellcolor{grpGeneral}51.3 & \cellcolor{grpGeneral}60.8 & \cellcolor{grpGeneral}37.2 & \cellcolor{grpGeneral}78.1 & \cellcolor{grpGeneral}85.5 & \cellcolor{grpGeneral}68.2 & \cellcolor{grpOCR}75.7 & \cellcolor{grpOCR}67.7 & \cellcolor{grpOCR}70.9 & \cellcolor{grpOCR}66.0 & \cellcolor{grpChart}80.7 & \cellcolor{grpChart}75.5 & \cellcolor{grpSpatial}\textbf{73.9} & \cellcolor{grpSpatial}0.2 & \cellcolor{grpCaption}30.4 & \cellcolor{grpCaption}12.2 & \cellcolor{grpMath}24.4 & \cellcolor{grpAvg}57.9 \\
        InternVL3.5-8B \hfill $\diamondsuit$            & \cellcolor{grpGeneral}2217 & \cellcolor{grpGeneral}\textbf{53.7} & \cellcolor{grpGeneral}62.4 & \cellcolor{grpGeneral}\textbf{38.0} & \cellcolor{grpGeneral}76.2 & \cellcolor{grpGeneral}86.3 & \cellcolor{grpGeneral}62.9 & \cellcolor{grpOCR}72.6 & \cellcolor{grpOCR}62.3 & \cellcolor{grpOCR}66.4 & \cellcolor{grpOCR}58.4 & \cellcolor{grpChart}79.0 & \cellcolor{grpChart}74.8 & \cellcolor{grpSpatial}69.9 & \cellcolor{grpSpatial}39.2 & \cellcolor{grpCaption}29.9 & \cellcolor{grpCaption}5.6 & \cellcolor{grpMath}38.1 & \cellcolor{grpAvg}58.6 \\
        Nemotron-Nano-12B \hfill $\diamondsuit$         & \cellcolor{grpGeneral}2004 & \cellcolor{grpGeneral}50.7 & \cellcolor{grpGeneral}59.9 & \cellcolor{grpGeneral}36.3 & \cellcolor{grpGeneral}77.9 & \cellcolor{grpGeneral}85.2 & \cellcolor{grpGeneral}69.9 & \cellcolor{grpOCR}\textbf{80.2} & \cellcolor{grpOCR}69.0 & \cellcolor{grpOCR}69.7 & \cellcolor{grpOCR}64.9 & \cellcolor{grpChart}\textbf{83.9} & \cellcolor{grpChart}78.6 & \cellcolor{grpSpatial}64.7 & \cellcolor{grpSpatial}62.3 & \cellcolor{grpCaption}18.8 & \cellcolor{grpCaption}2.9 & \cellcolor{grpMath}33.1 & \cellcolor{grpAvg}60.0 \\
        Qwen3.5-9B \hfill $\diamondsuit$                & \cellcolor{grpGeneral}2226 & \cellcolor{grpGeneral}52.2 & \cellcolor{grpGeneral}57.6 & \cellcolor{grpGeneral}36.7 & \cellcolor{grpGeneral}\textbf{78.5} & \cellcolor{grpGeneral}84.7 & \cellcolor{grpGeneral}\textbf{74.6} & \cellcolor{grpOCR}71.7 & \cellcolor{grpOCR}\textbf{70.8} & \cellcolor{grpOCR}\textbf{76.3} & \cellcolor{grpOCR}\textbf{71.6} & \cellcolor{grpChart}83.5 & \cellcolor{grpChart}\textbf{80.3} & \cellcolor{grpSpatial}69.3 & \cellcolor{grpSpatial}83.6 & \cellcolor{grpCaption}33.7 & \cellcolor{grpCaption}\textbf{13.5} & \cellcolor{grpMath}\textbf{54.4} & \cellcolor{grpAvg}\textbf{65.2} \\
        \midrule
        EuroVLM-9B \hfill $\diamondsuit\clubsuit$       & \cellcolor{grpGeneral}1581 & \cellcolor{grpGeneral}34.0 & \cellcolor{grpGeneral}40.3 & \cellcolor{grpGeneral}19.8 & \cellcolor{grpGeneral}66.8 & \cellcolor{grpGeneral}84.2 & \cellcolor{grpGeneral}55.2 & \cellcolor{grpOCR}46.4 & \cellcolor{grpOCR}54.1 & \cellcolor{grpOCR}56.4 & \cellcolor{grpOCR}34.0 & \cellcolor{grpChart}61.5 & \cellcolor{grpChart}47.6 & \cellcolor{grpSpatial}38.8 & \cellcolor{grpSpatial}5.1 & \cellcolor{grpCaption}23.9 & \cellcolor{grpCaption}5.3 & \cellcolor{grpMath}12.9 & \cellcolor{grpAvg}41.3 \\
        TowerVision-9B \hfill $\diamondsuit\heartsuit$  & \cellcolor{grpGeneral}1414 & \cellcolor{grpGeneral}39.1 & \cellcolor{grpGeneral}41.9 & \cellcolor{grpGeneral}21.7 & \cellcolor{grpGeneral}70.8 & \cellcolor{grpGeneral}85.1 & \cellcolor{grpGeneral}49.9 & \cellcolor{grpOCR}47.5 & \cellcolor{grpOCR}58.3 & \cellcolor{grpOCR}56.1 & \cellcolor{grpOCR}38.3 & \cellcolor{grpChart}65.0 & \cellcolor{grpChart}39.2 & \cellcolor{grpSpatial}49.1 & \cellcolor{grpSpatial}8.3 & \cellcolor{grpCaption}24.4 & \cellcolor{grpCaption}4.9 & \cellcolor{grpMath}14.6 & \cellcolor{grpAvg}42.5 \\
        Molmo2-8B \hfill $\diamondsuit\heartsuit$ & \cellcolor{grpGeneral}1936 & \cellcolor{grpGeneral}52.7 & \cellcolor{grpGeneral}61.2 & \cellcolor{grpGeneral}36.4 & \cellcolor{grpGeneral}73.0 & \cellcolor{grpGeneral}87.2 & \cellcolor{grpGeneral}66.8 & \cellcolor{grpOCR}59.0 & \cellcolor{grpOCR}61.2 & \cellcolor{grpOCR}62.8 & \cellcolor{grpOCR}50.2 & \cellcolor{grpChart}80.9 & \cellcolor{grpChart}43.5 & \cellcolor{grpSpatial}62.1 & \cellcolor{grpSpatial}3.6 & \cellcolor{grpCaption}19.5 & \cellcolor{grpCaption}3.8 & \cellcolor{grpMath}22.4 & \cellcolor{grpAvg}50.8 \\

        LLaVA-OV1.5-8B \hfill $\diamondsuit\heartsuit$  & \cellcolor{grpGeneral}2244 & \cellcolor{grpGeneral}52.9 & \cellcolor{grpGeneral}\textbf{64.7} & \cellcolor{grpGeneral}36.9 & \cellcolor{grpGeneral}75.8 & \cellcolor{grpGeneral}86.4 & \cellcolor{grpGeneral}62.9 & \cellcolor{grpOCR}76.1 & \cellcolor{grpOCR}63.3 & \cellcolor{grpOCR}72.0 & \cellcolor{grpOCR}62.6 & \cellcolor{grpChart}81.6 & \cellcolor{grpChart}74.2 & \cellcolor{grpSpatial}60.2 & \cellcolor{grpSpatial}78.1 & \cellcolor{grpCaption}36.5 & \cellcolor{grpCaption}8.2 & \cellcolor{grpMath}23.1 & \cellcolor{grpAvg}60.9 \\
        \midrule
        Salamandra-VL-7B \hfill $\diamondsuit\heartsuit\clubsuit$ & \cellcolor{grpGeneral}784 & \cellcolor{grpGeneral}27.7 & \cellcolor{grpGeneral}32.0 & \cellcolor{grpGeneral}11.8 & \cellcolor{grpGeneral}33.3 & \cellcolor{grpGeneral}64.9 & \cellcolor{grpGeneral}19.7 & \cellcolor{grpOCR}21.0 & \cellcolor{grpOCR}1.6 & \cellcolor{grpOCR}6.0 & \cellcolor{grpOCR}8.7 & \cellcolor{grpChart}30.1 & \cellcolor{grpChart}14.0 & \cellcolor{grpSpatial}26.3 & \cellcolor{grpSpatial}0.0 & \cellcolor{grpCaption}19.7 & \cellcolor{grpCaption}8.2 & \cellcolor{grpMath}3.1 & \cellcolor{grpAvg}19.8 \\
        Molmo2-O-7B \hfill $\diamondsuit\heartsuit\clubsuit$ & \cellcolor{grpGeneral}1618 & \cellcolor{grpGeneral}37.8 & \cellcolor{grpGeneral}52.6 & \cellcolor{grpGeneral}26.5 & \cellcolor{grpGeneral}70.0 & \cellcolor{grpGeneral}81.5 & \cellcolor{grpGeneral}58.3 & \cellcolor{grpOCR}57.4 & \cellcolor{grpOCR}55.1 & \cellcolor{grpOCR}52.7 & \cellcolor{grpOCR}41.6 & \cellcolor{grpChart}66.4 & \cellcolor{grpChart}40.4 & \cellcolor{grpSpatial}54.7 & \cellcolor{grpSpatial}3.4 & \cellcolor{grpCaption}18.4 & \cellcolor{grpCaption}4.7 & \cellcolor{grpMath}11.8 & \cellcolor{grpAvg}43.9 \\
        Perception-LM-8B \hfill $\diamondsuit\heartsuit\clubsuit$ & \cellcolor{grpGeneral}1368 & \cellcolor{grpGeneral}37.7 & \cellcolor{grpGeneral}\underline{52.8} & \cellcolor{grpGeneral}22.9 & \cellcolor{grpGeneral}73.0 & \cellcolor{grpGeneral}82.8 & \cellcolor{grpGeneral}53.4 & \cellcolor{grpOCR}\underline{76.5} & \cellcolor{grpOCR}60.8 & \cellcolor{grpOCR}65.4 & \cellcolor{grpOCR}\underline{60.3} & \cellcolor{grpChart}67.6 & \cellcolor{grpChart}\underline{69.2} & \cellcolor{grpSpatial}\underline{62.2} & \cellcolor{grpSpatial}0.0 & \cellcolor{grpCaption}15.2 & \cellcolor{grpCaption}3.1 & \cellcolor{grpMath}4.7 & \cellcolor{grpAvg}47.6 \\

        \textbf{\modelname \hfill $\diamondsuit\heartsuit\clubsuit$} & \cellcolor{grpGeneral}1815 & \cellcolor{grpGeneral}40.0 & \cellcolor{grpGeneral}44.7 & \cellcolor{grpGeneral}\underline{27.9} & \cellcolor{grpGeneral}72.0 & \cellcolor{grpGeneral}\underline{\textbf{89.3}} & \cellcolor{grpGeneral}\underline{60.9} & \cellcolor{grpOCR}59.0 & \cellcolor{grpOCR}66.2 & \cellcolor{grpOCR}68.7 & \cellcolor{grpOCR}45.0 & \cellcolor{grpChart}68.4 & \cellcolor{grpChart}67.8 & \cellcolor{grpSpatial}53.7 & \cellcolor{grpSpatial}77.4 & \cellcolor{grpCaption}\underline{\textbf{50.8}} & \cellcolor{grpCaption}\underline{11.2} & \cellcolor{grpMath}12.2 & \cellcolor{grpAvg}54.5 \\
        \textbf{\modelname-DPO \hfill $\diamondsuit\heartsuit\clubsuit$} & \cellcolor{grpGeneral}\underline{1824} & \cellcolor{grpGeneral}\underline{40.3} & \cellcolor{grpGeneral}44.0 & \cellcolor{grpGeneral}27.1 & \cellcolor{grpGeneral}\underline{73.8} & \cellcolor{grpGeneral}89.1 & \cellcolor{grpGeneral}59.9 & \cellcolor{grpOCR}62.4 & \cellcolor{grpOCR}\underline{68.0} & \cellcolor{grpOCR}\underline{69.5} & \cellcolor{grpOCR}45.6 & \cellcolor{grpChart}\underline{69.3} & \cellcolor{grpChart}66.5 & \cellcolor{grpSpatial}57.0 & \cellcolor{grpSpatial}\underline{79.2} & \cellcolor{grpCaption}49.7 & \cellcolor{grpCaption}10.7 & \cellcolor{grpMath}\underline{14.0} & \cellcolor{grpAvg}\underline{55.1} \\
        \bottomrule
    \end{tabular}%
    }
    \raggedright
\tiny{Notation: $^\dagger$denotes held-in benchmarks. 
$\diamondsuit$~Open weights; $\heartsuit$~Fully open vision data; $\clubsuit$~Fully open LLM.
\\ \textbf{Bold} = best overall, \underline{underline} = best fully open ($\diamondsuit$$\heartsuit$$\clubsuit$). }

\end{table*}

\subsection{Visual Instruction Following}

In Table~\ref{tab:main_results_portuguese}, we show the results of evaluating \modelname{} on a wide variety of benchmarks in pt-PT. Overall, these results show that \modelname{} sets a competitive baseline against fully open models with particularly strong results in captioning, spatial grounding, and OCR tasks. In the latter, it achieves best in-class performance in 5 out of 8 benchmarks in these categories, highlighting the benefit of the 6 purposely-built datasets constructed for these specific task-types. Complex visual reasoning emerges as a challenge with more limited performance in tasks such as MathVision and MMStar.
Focusing on baseline performance, interestingly, European-centric models, TowerVision and EuroVLM, do not demonstrate any advantage over similarly open peers. This indicates that general European language support does not inherently translate to robust pt-PT performance.

\vspace{-2mm}
\subsubsection{General VQA} category measures \modelname{}'s overall VQA capabilities across a wide variety of tasks, including logical reasoning, fine-grained VQA, basic OCR, code comprehension, scene understanding, and pop-culture knowledge. This inherent task diversity leads to high variance in performance for both \modelname{} and the baseline models. In this category \modelname{} demonstrates very competitive performance, posting the best scores in 6 out 7 benchmarks among fully open models, but we also observe that this is one of the categories with the biggest score disparity between fully open and just open weights models. Here, DPO seems to have mixed impact with minor improvements and regressions. 
These results ultimately suggest that General VQA performance is still strongly bounded by the availability of licensable broad world-knowledge training data, reflected in the gap between the best fully open (AMALIA) and best open weights (Qwen3.5) models.

\vspace{-2mm}
\subsubsection{Optical Character Recognition (OCR)} presents a unique evaluation challenge, as many input images contain mostly English text and our evaluation prompts are in pt-PT. 
We observed that this language change occasionally causes \modelname{} to erroneously translate the extracted text into pt-PT rather than performing direct text extraction (e.g. the target answer is "Boat" and the model answers "Barco"). 
The current performance is the result of translating, during development, some OCR training datasets to replicate this setting, which led to a significant improvement in OCR performance in the validation benchmark set. Here, DPO has a positive impact, improving all benchmarks, as it benefits the most benchmarks affected by response style and format compliance.

\vspace{-2mm}
\subsubsection{Charts and Diagrams} shows that \modelname{} has again very competitive performance in chart comprehension tasks and biology knowledge, being the second best open model, on average, with particularly competitive performance in AI2D and closely trailing Perception-LM on ChartQA. 

\vspace{-2mm}
\subsubsection{Spatial Understanding.}

\begin{table*}[t]
\caption{RefCOCO Rec pt-PT results across different bounding box formats. $^*$~Indicates the default benchmark configuration.}
\label{tab:refcoco_ablation}
    \centering
    \small
    \resizebox{\textwidth}{!}{%
    \begin{tabular}{l *{12}{r}}
    \toprule
    & \multicolumn{2}{c}{\textbf{Gemma3}} & \multicolumn{2}{c}{\textbf{GLM4v}} & \multicolumn{2}{c}{\textbf{Qwen3vl}} & \multicolumn{2}{c}{\textbf{Qwen3.5}} & \multicolumn{2}{c}{\textbf{Molmo2-8B}} & \multicolumn{2}{c}{\textbf{\modelname{-DPO}}} \\
    \cmidrule(lr){2-3} \cmidrule(lr){4-5} \cmidrule(lr){6-7} \cmidrule(lr){8-9} \cmidrule(lr){10-11} \cmidrule(lr){12-13} 
    \textbf{Format} & [0-1] & [0-1000] & [0-1] & [0-1000] & [0-1] & [0-1000] & [0-1] & [0-1000] & [0-1] & [0-1000] & [0-1] & [0-1000] \\
    \midrule
    $[x_{min}, y_{min}, w, h]$ & \ha{6.3} & \ha{1.1} & \ha{0.0} & \ha{24.7} & \ha{0.2} & \ha{25.05} & \ha{6.6} & \ha{23.9} & \ha{2.9} & \ha{4.1} & \ha{23.3} & \ha{25.9} \\
    $[x_{min}, y_{min}, x_{max}, y_{max}]$ & $^*$\ha{6.1} & \ha{3.6} & $^*$\ha{86.9} & \ha{87.3} & $^*$\ha{0.2} & \ha{88.3} & $^*$\ha{83.6} & \ha{86.4} & $^*$\ha{3.6} & \ha{4.2} & $^*$\ha{79.2} & \ha{75.4} \\
    \bottomrule
    \end{tabular}
    }
\end{table*}

Spatial understanding tasks require the model to have a strong grasp of what is present in each image region.
The RefCOCO Rec~\cite{refcoco} benchmark, a bounding box prediction task, emerges as an outlier task with a pronounced performance range, particularly affecting open-source models. Closer examination of outputs reveals a systematic struggle by low-performing models to adhere to the bounding box format specified by the task. Table~\ref{tab:refcoco_ablation} shows an ablation study using several formats and scales. The results highlight that most models lack format flexibility when prompted in pt-PT and often fail to adapt, while \modelname{} demonstrates adaptability in varying specifications. DPO further solidifies \modelname{}'s strong spatial understanding, as both benchmarks see an improvement, particularly EmbSp. that gains over 3 points.

\vspace{-2mm}
\subsubsection{Captioning} emerges as the task-type in which \modelname{} demonstrates its strongest performance. This advantage stems from the task's requirement for syntactically correct pt-PT descriptions, in contrast to the short-form answers evaluated by most other benchmarks. \modelname{}'s success in these tasks highlights the need for native pt-PT training, demonstrating that the performance advantage exhibited by other models dwindles when long-form pt-PT generation is required.

\vspace{-2mm}
\subsubsection{MathVision} requires open-ended mathematical reasoning, making it exceptionally challenging. Here, \modelname{} has the best performance among open models but, similarly to General VQA, we note a significant performance gap in pt-PT MathVision against open weight s models. We believe the open model's performance stems from a lack of reasoning capability, specifically for \modelname{} we lack long-form pt-PT reasoning data in our training mix, which limits its ability to solve complex visual math problems. Furthermore, the baseline results, where most models score below 20, highlight a broader gap in pt-PT reasoning capabilities.

\section{Conclusion}

This paper introduces \modelname{}, the first LVLM that natively targets the European Portuguese language variety, a challenging low-resource setting lacking public datasets, models, or benchmarks.
We addressed cross-dialect leakage from pt-BR and the absence of native pt-PT multimodal resources by leveraging open-source English datasets coupled with machine translation using validated models. We further complemented this with complex synthetic dataset generation pipelines that target specific model behaviours absent from the collected data.
To train \modelname{}, we employed a three-stage process that spans the entire LVLM training cycle: modality alignment, visual instruction tuning, and preference optimization.
Furthermore, to address the lack of benchmarks, we translated 18 SoTA multimodal benchmarks to pt-PT.
Comprehensive evaluation of \modelname{} against 15 baselines shows it is extremely competitive among open-source models in European Portuguese, top scoring in 13 out 18 benchmarks, particularly excelling in captioning, spatial grounding, and OCR, tasks that demand strong pt-PT textual comprehension paired with fine-grained image understanding.
With this work, we opened the door for future pt-PT LVLM research as we provided resources for the full cycle of model development.

\section{Acknowledgements}
This work was supported by the AMALIA project under Measure RE-C05-i08 of the Portuguese national ``Programa de Recuperação e Resiliência''. We also acknowledge the support of Fundação para a Ciência e Tecnologia (FCT) and the NOVA LINCS project (UID/04516/2025). We thank the Barcelona Supercomputing Center (BSC) for providing the computational resources that made this work possible.
We also thank Professor André Martins for his valuable feedback.

\bibliographystyle{splncs04}
\bibliography{bibliography}

\end{document}